\documentclass[10pt,twocolumn,letterpaper]{article}

\usepackage[pagenumbers]{cvpr} 

\usepackage{graphicx}
\usepackage{times}
\usepackage{helvet}
\usepackage{courier}
\usepackage{amsmath}
\usepackage{algorithm}
\usepackage{algorithmic}
\usepackage{csquotes} 
\usepackage{color}
\usepackage{paralist}
\usepackage{amssymb}
\usepackage{indentfirst}
\usepackage{float}
\usepackage{multirow}
\usepackage{cite}
\usepackage{mathrsfs}

\usepackage{mathrsfs}
\usepackage{textcomp,booktabs}
\usepackage{amssymb}
\usepackage{pifont}
\usepackage[misc]{ifsym}
\newcommand{\cmark}{\ding{51}}%
\newcommand{\xmark}{\ding{55}}%

\usepackage{mathrsfs}
\usepackage{color}
\definecolor{SeaGreen4}{RGB}{0,205,102} 
\definecolor{SlateBlue}{RGB}{106,90,205} 
\definecolor{DarkRed}{RGB}{178,34,34} 
\usepackage[colorlinks, linkcolor=red,  anchorcolor=blue, citecolor=SlateBlue]{hyperref}

\usepackage{colortbl}
\definecolor{mygray}{gray}{.9}
\definecolor{mypink}{rgb}{.99,.91,.95}
\definecolor{mycyan}{cmyk}{.3,0,0,0}

\usepackage[capitalize]{cleveref}
\crefname{section}{Sec.}{Secs.}
\Crefname{section}{Section}{Sections}
\Crefname{table}{Table}{Tables}
\crefname{table}{Tab.}{Tabs.}

\usepackage[dvipsnames]{xcolor}

\title{ CRSOT: Cross-Resolution Object Tracking using Unaligned Frame and Event Cameras }

\author{Yabin Zhu$^{1,5}$, Xiao Wang$^{1,2}$\thanks{\Letter~~Corresponding author: Xiao Wang (xiaowang@ahu.edu.cn)}, 
	Chenglong Li$^{1,3}$, Bo Jiang$^{1,2}$, Lin Zhu$^{4}$, Zhixiang Huang$^{1,5}$, \\ Yonghong Tian$^{6,7,8}$, Jin Tang$^{1,2}$ \\ 
${^1}${Key Laboratory of Intelligent Computing and Signal Processing of Ministry of Education, 
					Anhui University} \\
${^2}${School of Computer Science and Technology, Anhui University, China} \\
${^3}${School of Artificial Intelligence, Anhui University,  China} \\
${^4}${Beijing Institute of Technology, Beijing, China} \\
${^5}${School of Electronic and Information Engineering, Anhui University,  China} \\
${^6}${Peng Cheng Laboratory, Shenzhen, China} \\ 
${^7}${School of Computer Science, Peking University, Beijing, China} \\ 
${^8}${School of Electronic and Computer Engineering, Peking University, Shenzhen, China} \\ 
}

\begin{document}
\maketitle

\begin{abstract}
Existing datasets for RGB-DVS tracking are collected with DVS346 camera and their resolution ($346 \times 260$) is low for practical applications. Actually, only visible cameras are deployed in many practical systems, and the newly designed neuromorphic cameras may have different resolutions. The latest neuromorphic sensors can output high-definition event streams, but it is very difficult to achieve strict alignment between events and frames on both spatial and temporal views. Therefore, how to achieve accurate tracking with unaligned neuromorphic and visible sensors is a valuable but unresearched problem. In this work, we formally propose the task of object tracking using unaligned neuromorphic and visible cameras. We build the first unaligned frame-event dataset CRSOT collected with a specially built data acquisition system, which contains 1,030 high-definition RGB-Event video pairs, 304,974 video frames. In addition, we propose a novel unaligned object tracking framework that can realize robust tracking even using the loosely aligned RGB-Event data. Specifically, we extract the template and search regions of RGB and Event data and feed them into a unified ViT backbone for feature embedding. Then, we propose uncertainty perception modules to encode the RGB and Event features, respectively, then, we propose a modality uncertainty fusion module to aggregate the two modalities. These three branches are jointly optimized in the training phase. Extensive experiments demonstrate that our tracker can collaborate the dual modalities for high-performance tracking even without strictly temporal and spatial alignment. The source code, dataset, and pre-trained models will be released at \textbf{\url{https://github.com/Event-AHU/Cross_Resolution_SOT}}. 
\end{abstract}

\section{Introduction} \label{sec:intro}
The target of visual tracking is to locate the specified target object smoothly by adjusting the location and scale of the bounding box. The performance under challenging scenarios (e.g., fast motion, illumination variation) is still unsatisfactory, evening strong and deep neural networks are utilized~\cite{mayer2022transforming, wang2021transformer, voigtlaender2020siam}. Most previous trackers are developed based on frame-based sensors, however, some researchers find that the poor performance is caused by the high latency of the imaging mechanism of RGB cameras \cite{kiani2017need, wang2019event, tulyakov2021time, wang2021viseventbenchmark, iccvFE108}. Therefore, they adopt the bio-inspired Dynamic Vision Sensors \cite{brandli2014dvs240, posch2010qvga, finateu20205, chen2019Celex5} (DVS, also called Event Camera) to handle the challenging tracking task in the wild \cite{wang2021viseventbenchmark, iccvFE108, zhang2021MEEvent, yang2019dashnet, chen2020end, chamorro2020high, chen2019asynchronous, huang2018event, gehrig2020eklt, camunas2017event}. The DVS has shown its advantages in many aspects compared with traditional RGB cameras, especially on the \emph{low latency, low power, high dynamic range,} and \emph{high temporal resolution} \cite{gallego2020eventSurvey}. More in detail, the DVS output asynchronous \emph{events}, and each event denotes the light changes outstrip the pre-defined threshold. The increase and decrease of light intensity of each pixel is denoted as ON and OFF event, respectively. Due to the unique imaging mechanism, the DVS is not good at capturing static objects or targets with very slow motion. Fortunately, the RGB camera works well in this situation and outputs video frames with helpful color and texture details. Therefore, it is natural to combine the RGB and DVS for high-performance tracking.

\begin{figure}
\center
\includegraphics[width=3.3in]{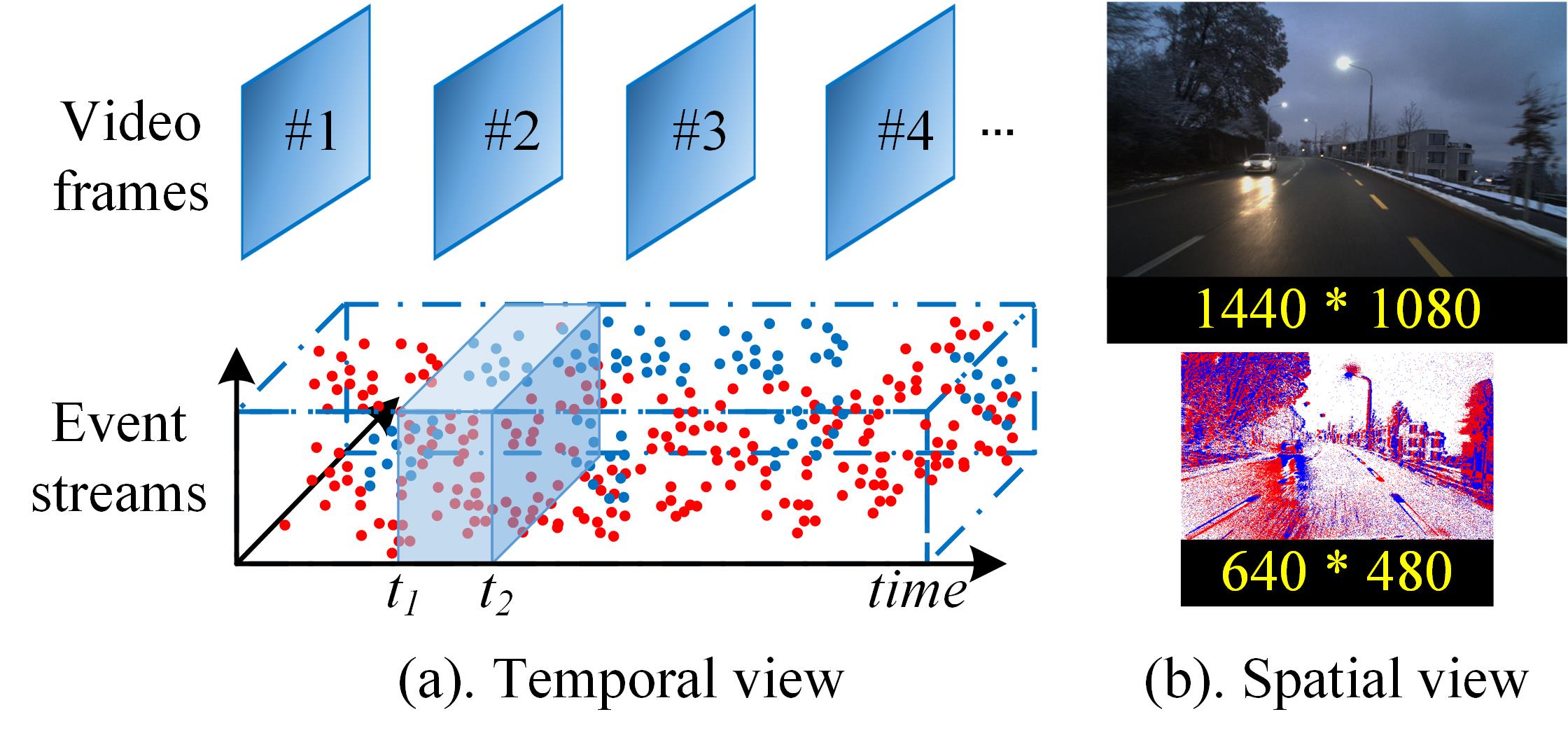}
\caption{Illustration of cross-resolution object tracking with unaligned video frames and event streams.} 
\label{frontIMGs}
\end{figure}

Recently, the FE108~\cite{iccvFE108}, VisEvent~\cite{wang2021viseventbenchmark}, and COESOT~\cite{tang2022COESOT} collected with a DVS346 camera are proposed for RGB-DVS tracking, however, their resolution is $346 \times 260$ which is relatively low for practical applications. Actually, only visible cameras are deployed in many practical systems. The newly designed neuromorphic cameras can output high-definition event streams but may have different resolutions, as shown in Fig.~\ref{frontIMGs}. It is very difficult to achieve strict alignment between event steams and RGB frames on both spatial and temporal views. Therefore, how to achieve high-performance tracking with unaligned neuromorphic and visible sensors is a valuable but unresearched problem.

In this paper, we formally propose the task of object tracking using unaligned neuromorphic and visible cameras. Specifically, we first build a new data acquisition system that contains the RGB frame camera ($1440 \times 1080$) and CeleX-V event camera ($1280 \times 800$). Then, we collect a large-scale, high-quality, and high-resolution benchmark dataset for this task, termed CRSOT. It contains 1,030 video sequences, 304,974 RGB frames, and these videos are split into training and testing subset which contains 836 and 194 videos, respectively. CRSOT covers a wide range of scenarios (e.g., indoor and outdoor, sunny day and raining weather), and challenging factors (e.g., fast motion, low illumination, background clutter). More details about our dataset can be found in Section~\ref{crsotnvDatset}. 

To build a more comprehensive benchmark, in this work, we also propose a new baseline for the unaligned RGB-Event tracking problem. Given the RGB frames and event streams, we first transform the continuous event streams into event images by stacking event points within a fixed time interval. Then, we resize the two modalities into the same resolution and adopt the ViT~\cite{dosovitskiy2020image} network with token elimination as a unified backbone for the feature extraction by following OSTrack~\cite{ye2022joint}. The template and search regions of dual modalities are extracted and fed into the backbone for feature extraction. To better handle the relaxed registration issue, in this work, we predict the probabilistic representation instead of regular point representation for RGB-Event based tracking. The template/search regions of RGB and event are fed into the MDUP module and CMDUP module for uncertainty perception by predicting its distribution representation via mean and variation. We also propose MUF (Modality Uncertainty Fusion) which can adaptively fuse RGB-Event feature representations. Finally, we feed the enhanced features into a tracking head which contains both classification and regression branches for target object localization. An overview of our proposed tracking framework can be found in Fig.~\ref{framework}.

To sum up, the contributions of this work can be concluded as follows: 

$\bullet$ We propose a new setting of object tracking with unaligned neuromorphic and visible cameras. It provides a new clue for introducing neuromorphic cameras into practical RGB camera-based monitoring systems. 

$\bullet$ We propose the first high-resolution, large-scale, and high-quality dataset for cross-resolution single object tracking using unaligned RGB-DVS cameras, termed CRSOT. It contains 1,030 RGB-DVS videos, 304,974 frames, and we split them into training and testing subsets with 836 and 194 videos, respectively. 

$\bullet$ We propose a novel unaligned object tracking framework that can realize robust tracking even using the unaligned RGB-Event data. 

Extensive experiments on multiple benchmark datasets demonstrate the effectiveness of our proposed framework. We hope this work can attract more researchers on the unaligned dual-modality tracking problem.

\section{Related Work} \label{relatework}

\noindent \textbf{RGB-DVS Tracking.~} 
Due to the robustness of DVS to the aforementioned challenges, some researchers have begun to utilize DVS for tracking. Specifically, a parametric object-level motion/transform model is learned for event-based tracking \cite{chen2020end}. Chen et al. \cite{chen2019asynchronous} propose an event-to-frame conversion algorithm, termed ATSLTD, and feed the ATSLTD frames into ETD method for tracking. 
Wang et al.~\cite{wang2023eventvot} propose a cross-modality/view knowledge distill framework to improve the training of event-based tracker by learning from multi-modal or multi-view data. 
e-TLD \cite{ramesh2018long} use the event-based detector to help track in long-term settings. 
There are also many works that focus on feature tracking using DVS sensor \cite{alzugaray2020haste, gehrig2020eklt}. However, tracking based on DVS only is not reliable, as it only captures the dynamic regions, e.g., the edge of a moving object, and is unable to perceive the static or slow-moving targets well.

To avoid issues caused by a single camera, it is intuitive to combine the two sensors for robust object tracking. 
For example, Huang et al. \cite{huang2018event} propose tracking by fusing RGB and CeleX sensors for candidate search region mining and model update with samples reconstructed from event flows. 
Zhang et al. \cite{iccvFE108} propose to enhance RGB and event features via self-/cross-domain attention schemes. 
Wang et al. \cite{wang2021viseventbenchmark} propose the Cross-Modality Transformers (CMT) to fuse the RGB and DVS features for tracking. 
Tang et al.~\cite{tang2022COESOT} propose a unified tracking backbone to achieve RGB-Event feature extraction and fusion simultaneously, termed CEUTrack. 
A mask modeling strategy is proposed by Zhu et al.~\cite{Zhu2023CrossmodalOH} which target to address the issue of cross-modal interaction between RGB and event data. 
DANet~\cite{Fu2023DistractorAwareET} proposed by Fu et al. aggregate the Transformer and Siamese architecture to achieve an event-based interference sensing tracking.  
ViPT~\cite{zhu2023visual} proposed by Zhu et al. incorporates learnable modal-relevant prompts while fixing the weights of pre-trained models which enhance the adaptability of the models to diverse multi-modal tracking tasks. 
Zhu et al.~\cite{Zhu2022LearningGK} process event clouds using the graph method and predict the motion-aware target likelihood for event-based tracking.  
A cross-domain attention fusion algorithm STNet~\cite{Zhang2022SpikingTF} is proposed by Zhang et al. which achieves good performance on event data. 
More questions still need to be solved for this task, such as how to design more suitable modal alignment modules and fusion modules in actual unaligned scenes.

\noindent \textbf{Neuromorphic Tracking Datasets.~} 
As it is a newly arising research topic, the DVS-based tracking datasets are significantly less than RGB-based benchmarks. Early researchers conduct their experiments using simulated datasets which are transformed or recorded based on off-the-shelf RGB-based tracking datasets. For example, 
Hu et al. \cite{hu2016dvs} adopt the DAViS240C sensor to get events at a resolution of $240 \times 180$ by recording the screen. 
Huang et al. \cite{huang2018event} also use the CeleX camera to get the events of RGB videos. Obviously, these datasets maybe can't fully reflect real challenges in the real world. 
Liu et al. \cite{liu2016combined} record a real event dataset Ulster, but only contains one video sequence. 
EED \cite{mitrokhin2018event} was proposed in 2018, but it also only has 7 video pairs. 
Zhang et al. propose a new dataset that contains 108 videos, termed FE108 \cite{iccvFE108}, but this dataset is almost saturated, as the baseline method already achieves $92.4\%$ on the precision plot. 
Wang et al. contribute a VisEvent \cite{wang2021viseventbenchmark} benchmark dataset which contains 820 videos and multiple baselines. 
Tang et al. propose a new dataset termed COESOT~\cite{tang2022COESOT} which is category-wide and large-scale for this research area. 
However, the resolution ($346 \times 260$) of these datasets is limited due to the use of DVS346 sensors. 
These datasets cannot meet certain scenarios that require high-definition resolution, for example, military filed and autonomous vehicles. In contrast, our newly proposed CRSOT is a high-resolution, high-quality, and large-scale frame-event tracking dataset. We believe this dataset will provide a good platform for trackers to evaluate unaligned high-resolution RGB-DVS videos.

\noindent \textbf{Uncertainty-aware Learning.}  
Unlike previous point embedding representations, uncertainty learning is a probabilistic distribution representation that improves the robustness and generalization ability of the network through diverse inference. It has been widely used in many vision tasks such as face recognition, object detection, cross-modal matching, and multi-modal fusion. 
Specifically, Shi et al.~\cite{shi2019probabilistic} introduce uncertainty learning for the first time by modeling face image embedding as Gaussian distributions to account for uncertainty. 
Chang et al.~\cite{chang2020data} propose a method based on~\cite{shi2019probabilistic} that simultaneously learns the mean and variance of the features to model the Gaussian distribution, thereby achieving more robust performance on low-quality face datasets. 
Li et al.~\cite{li2022uncertainty} adapt distance-aware uncertainty estimation to solve unknown object detection tasks. 
Ji et al.~\cite{ji2023map} introduce uncertainty in vision-language contrastive learning, masked language modeling and image-text matching, which solves the problem of understanding the multi-modal uncertainty correspondences. 
Zhang et al.~\cite{zhang2023provable} clarify the relationship between uncertainty estimation and multimodal fusion and provide a theoretical foundation for multi-modal fusion with uncertainty. 
Inspired by these works, in this work, we propose a novel uncertainty-aware RGB-Event fusion framework that achieves high-performance tracking on various datasets.

\section{Tracking with Unaligned Frames and Events}  

\noindent \textbf{Problem Formulation.}   
Given the RGB frames $\mathcal{F} = [F_1, F_2, ..., F_N]$ and Event flows $\mathcal{E} = \{e_j\}_{j=1}^{T} = \{[x_j, y_j, p_j, t_j]\}_{j=1}^{T}$, where $F_i$ ($i =\{1, ..., N\}$) is the video frame, $N$ is the number of frames in the current video; $e_j$ ($j =\{1, ..., T\}$) is one event (or spike) of the event flow, $T$ is the total number of events, $x_j$ and $y_j$ are the coordinates, $t_j$ is the timestamp, $p_j \in \{+1, -1\}$ is the polarity which denotes the increased or decreased light intensity using $+1$ and $-1$ (also termed ON and OFF event), respectively. 
The goal of RGB-DVS tracking is to jointly utilize the two domains for more accurate and efficient tracking. Formally, we input the two data into the RGB-DVS tracker and output the trajectory of the initialized target object: 
\begin{equation}
\label{PFtrack} 
\{ [x^i, y^i, w^i, h^i] \}_{i=1}^{N} = Tracker ([\mathcal{F}, \mathcal{E}]), 
\end{equation}
where $[x^i, y^i, w^i, h^i]$ are the top-left coordinates, width, and height of the bounding box of frame $i$, respectively. The evaluation of tracking performance is conducted based on discrete video frames.

\begin{table*}
\center
\scriptsize    
\caption{\textbf{Comparison of existing event datasets for object tracking.} $\#$ denotes the number of corresponding items. Att, HR, and DW are short for Attributes, High Resolution, and Different Weathers. NIR means that the corresponding dataset is annotated under the guidance of near-infrared camera. 
} 
\label{benchmarkList}
\begin{tabular}{r|ccccccccccccccc}
\hline \toprule [0.5 pt]
\textbf{Datasets}    &\textbf{Year}	&\textbf{Project} &\textbf{\#Videos}  &\textbf{\#Frames}  &\textbf{\#Resolution}  &\textbf{\#Att} &\textbf{Aim} &\textbf{Absent} &\textbf{Real} &\textbf{Public}  &\textbf{Color}  &\textbf{HR}   &\textbf{DW}   &\textbf{NIR} \\ 
\hline
\textbf{VOT-DVS}~\cite{hu2016dvs}     &2016    & \href{https://dgyblog.com/projects-term/dvs-dataset.html}{URL} 	&60           &-   &$240 \times 180$  	 &-   &Eval   &\xmark   &\xmark     &\cmark &\xmark   &\xmark &\xmark &\xmark \\
\textbf{TD-DVS}~\cite{hu2016dvs}        &2016     & \href{https://dgyblog.com/projects-term/dvs-dataset.html}{URL}	&77          &-   &$240 \times 180$  	 &-   &Eval  &\xmark    &\xmark     &\cmark  &\xmark  &\xmark  &\xmark  &\xmark \\
\textbf{Ulster}~\cite{liu2016combined}   &2016      &-		&1     &9,000  		 &$240 \times 180$  	 &-   &Eval  &\xmark  &\cmark     &\xmark &\xmark    	&\xmark	&\xmark &\xmark \\
\textbf{EED}~\cite{mitrokhin2018event} &2018     & \href{http://prg.cs.umd.edu/BetterFlow.html}{URL}	&7     &234  &$240 \times 180$  	 &-   &Eval  &\xmark    &\cmark     &\cmark   &\xmark  &\xmark &\xmark &\xmark \\
\hline 
\textbf{FE108}~\cite{iccvFE108}	&2021     &\href{https://zhangjiqing.com/dataset/}{URL}		&108     &208,672   &$346 \times 260$  	 &-   &Train/Eval      &\xmark     &\cmark   &\cmark &\xmark   &\xmark  &\xmark &\xmark \\
\textbf{VisEvent}~\cite{wang2021viseventbenchmark}  	&2021     &\href{https://github.com/wangxiao5791509/VisEvent_SOT_Benchmark}{URL} &820       &371,127      	 &$346 \times 260$ 	&17     &Train/Eval &\cmark     &\cmark    &\cmark &\cmark   &\xmark  &\xmark &\xmark \\ 
\textbf{COESOT}~\cite{tang2022COESOT} &2022     &\href{https://github.com/Event-AHU/COESOT}{URL} &\textbf{1,354}   &478,721  &\textbf{$346 \times 260$} 	&17      &Train/Eval  &\cmark     &\cmark    &\cmark &\cmark   &\xmark  &\xmark &\xmark \\ 
\textbf{EventVOT}~\cite{wang2023eventvot} &2023     &\href{https://github.com/Event-AHU/EventVOT_Benchmark}{URL} &1,141   &\textbf{569,359}  &\textbf{$1280 \times 720$} 	&14      &Train/Eval   &\cmark     &\cmark    &\cmark &\xmark   &\cmark  &\xmark &\xmark \\ 
\hline 
\textbf{CRSOT (Ours)} &2023     &\href{https://sites.google.com/view/vdthr/}{URL} &1,030   & 304,974 &\textbf{1280} $\times$ \textbf{800} 	&17      &Train/Eval   &\cmark     &\cmark    &\cmark &\cmark   &\cmark  &\cmark &\cmark \\
\hline \toprule [0.5 pt]
\end{tabular}
\end{table*}

\begin{figure}[!htp]
    \centering
    \includegraphics[width=0.5\linewidth]{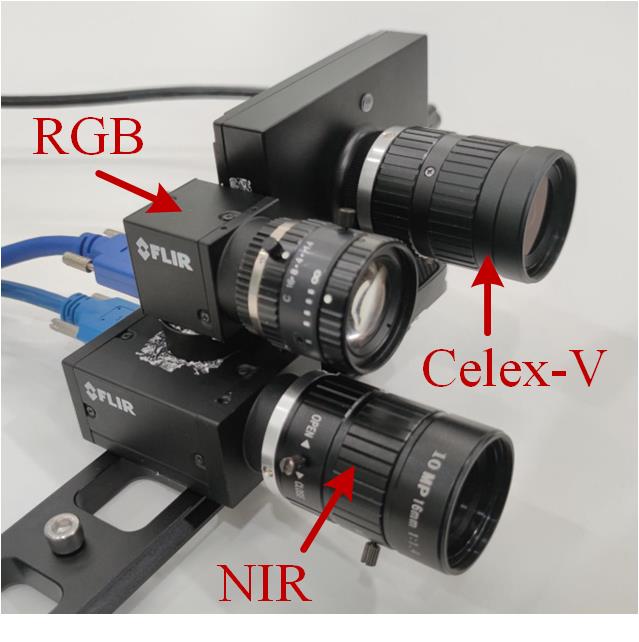}
    \caption{The designed camera system for data collection.} 
    \label{fig:cameraSYS}
\end{figure}

\noindent \textbf{Key Challenges.} Different from existing tracking datasets \cite{iccvFE108, wang2021viseventbenchmark, tang2022COESOT} which are aligned well on the hardware, our data acquisition device roughly click the record and stop button with a Python script. Thus, the time stamps of our RGB frames are not strictly aligned with the event flow. This will make the content of captured dual-modalities with slight differences. As these cameras have various resolutions, this problem will be further magnified. Therefore, how to conduct tracking on one modality (for example, the RGB frames) by referencing another one (i.e., the event streams, correspondingly), but without pixel-level alignment, is the key research point for cross-resolution object tracking. In addition, how to represent and learn the features of event streams is another problem worthy of study.

\begin{figure*}
\center
\includegraphics[width=7in]{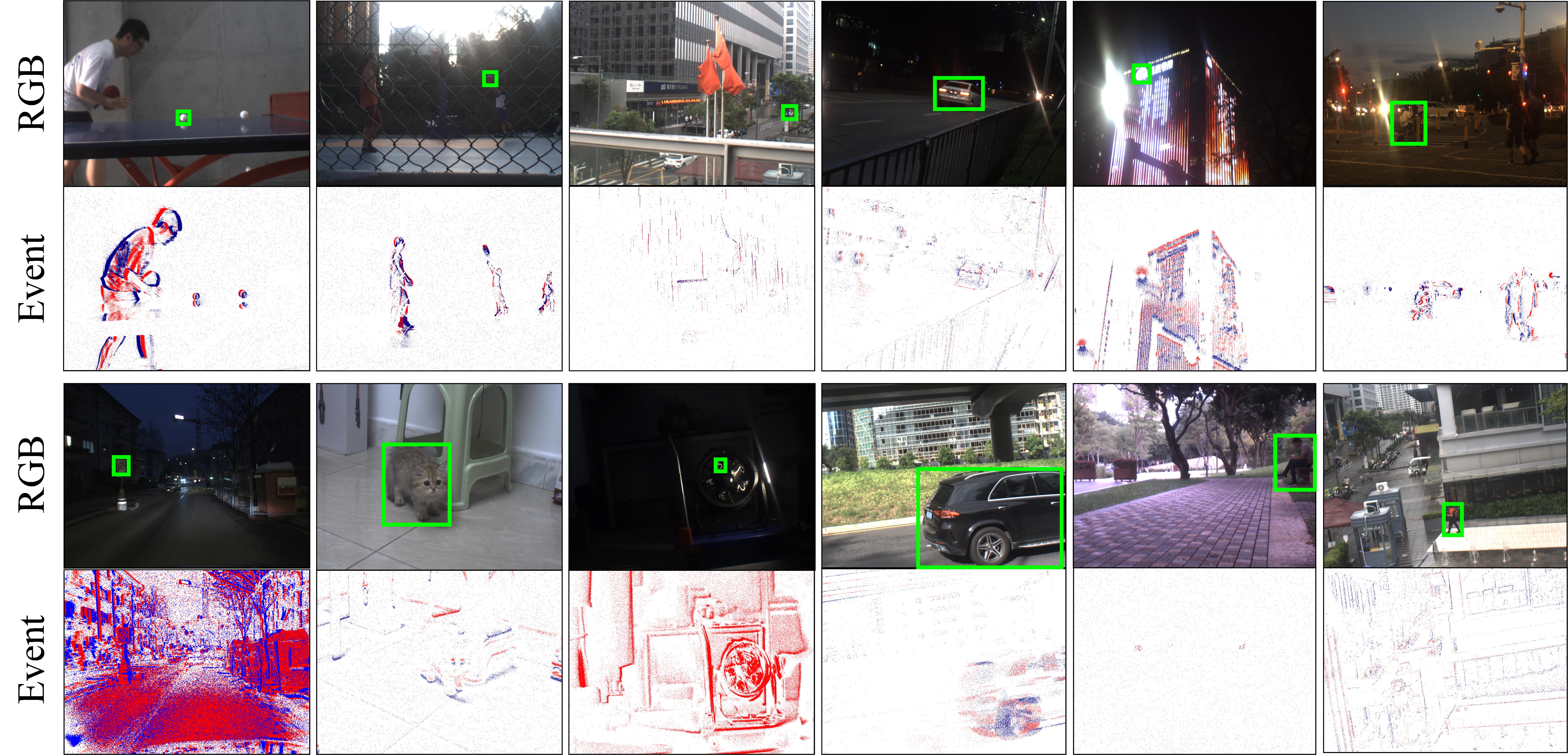}
\caption{Illustration of representative samples of our newly proposed CRSOT dataset. The resolution of dual modalities is resized for better visualization.}  
\label{CRSOT_benchmark}
\end{figure*}

\section{CRSOT Benchmark Dataset} \label{crsotnvDatset}

\subsection{Dataset Collection}

\noindent \textbf{Data Acquisition System.~} 
To acquire a high-resolution RGB-DVS tracking dataset, we built a hybrid camera system that contains three sensors, i.e., the CeleX-V, RGB, and NIR cameras. The default resolution of RGB and CeleX-V sensors are $1440 \times 1080$, and $1280 \times 800$, which are significantly better than DVS346 sensors ($346 \times 260$). The NIR camera is used to guide the annotation in the dark night which will make our ground truth more accurate. It is important for the annotation of videos in the degraded scene, especially at night time, however, this point is usually ignored by previous RGB-DVS tracking datasets. To make these cameras synchronized in time,  we wrote a recording software that can simultaneously trigger for recording. To make our dataset cover more scenarios, we borrow some videos from the DSEC dataset \cite{Gehrig21DSEC}. These videos are also recorded in real scenarios, but built for other tasks. Some samples of our CRSOT dataset are visualized in Fig. \ref{CRSOT_benchmark}. 
The images of our data acquisition system and more examples of our dataset can be found in Fig.~\ref{fig:cameraSYS}.

\noindent \textbf{Scene Selection and Features.~} 
To construct a large-scale and comprehensive RGB-DVS tracking dataset, the selection of \emph{shooting location} and \emph{target object} are the key factors. For the tracking scenarios, we select the home scenes, laboratory, gymnasium, inside of the vehicle, street, zoo, market, lake, UAV test site, etc. Therefore, we can capture diverse target objects, including articles for daily use (e.g., cup, phone), pedestrians, cars, basketball, badminton, ping-pong, tennis balls, animals (e.g., cats, monkeys, birds), boats, and UAVs. 
Our CRSOT also considers different weather conditions, such as fine-, cloudy-, and rainy-day. 
More importantly, the collected videos fully reflect the key features of DVS and also the popular challenging factors in the tracking task, such as high speed, low light, and cluttered background. 
Thanks to the NIR camera, we can also collect some videos in the dark night and the annotation problem in the low illumination can be greatly mitigated.

\subsection{Attribute Definition and Statistic Analysis}   
To evaluate the performance of trackers under each challenging factor, in this work, we define 17 attributes for the CRSOT dataset, including the motion of target object or cameras, i.e., CM (Camera Motion), ROT (Rotation), MB (Motion Blur), FM (Fast Motion), NM (No motion); illumination related attributes like OE (Over-Exposure), LI (Low Illumination); occlusion related attributes like FOC (Full Occlusion), POC (Partial Occlusion), etc. The complete list of these attributes can be found in Table~\ref{AttributeList}.

\begin{table}
\center
\small  
\caption{ The 17 attributes defined in our proposed CRSOT dataset.} 
\label{AttributeList}
\resizebox{0.48\textwidth}{!}{ 
\begin{tabular}{l|l} 
\hline 
\textbf{Attributes}    &\textbf{Description}  \\ 
\hline
01. CM (Camera Motion)       & The camera is moving when recording the videos \\ 
02. ROT (Rotation)           & The target object changes its views significantly \\ 
03. DEF (Deformation)        & The shape of target object changed  \\
04. FOC (Full Occlusion)     & The target object is fully occluded by other objects     \\
05. LI (Low Illumination)    & The target object is recorded in low illumination scenarios     \\
06. OV (Out-of-View)         & The target object moves out of the view of camera  \\ 
07. POC (Partial Occlusion)     &Part of target object is occluded     \\
08. VC (Viewpoint Change)       &The views of target object vary during tracking     \\
09. SV (Scale Variation)        &The width and height of target object changed significantly     \\
10. BC (Background Clutter)     &The target object is heavily influenced by background     \\
11. MB (Motion Blur)                &The imaging picture seems blur due to fast motion \\ 
12. ARC (Aspect Ration Change)      &The ratio of bounding box aspect ratio varied significantly     \\
13. FM (Fast Motion)                &Target object moves quickly  \\ 
14. NM (No motion)                  &The target object is stationary     \\
15. IV (Illumination Variation)     &The light intensity changes during tracking     \\
16. OE (Over-Exposure)              &The light intensity is very high     \\
17. BOM (Background Object Motion)  &The target object is heavily influenced by background     \\
\hline 
\end{tabular}
} 
\end{table}

\begin{figure}
\center
\includegraphics[width=\columnwidth]{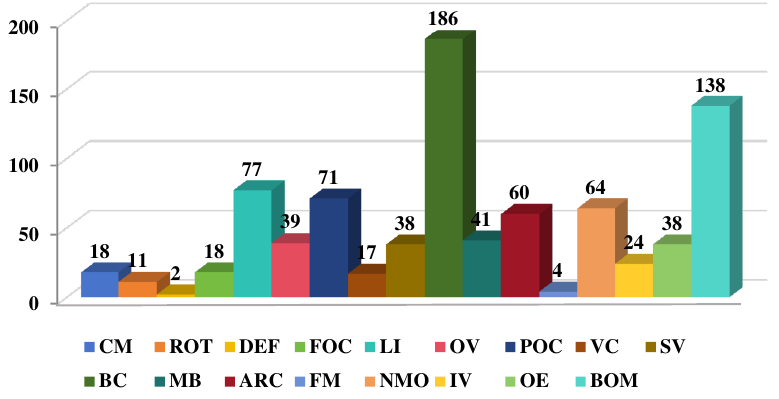}
\caption{Distribution of attributes defined on CRSOT testing set.} 
\label{attrDIST}
\end{figure}

From a statistical point of view, the proposed CRSOT contains 1,030 RGB-Event video pairs, 304,974 RGB frames. 
We split them into the training and testing subset which contains 836 and 194 videos, respectively. 
For the distribution of attributes defined on the CRSOT testing subset, as shown in Fig.~\ref{attrDIST}, we can find that most of the videos contain the challenge of BC (Background Clutter, 186 videos), BOM (Background Object Motion,  138 videos), LI (Low Illumination, 77 videos), POC (Partial Occlusion, 71 videos).

\begin{figure*}
\center
\includegraphics[width=7in]{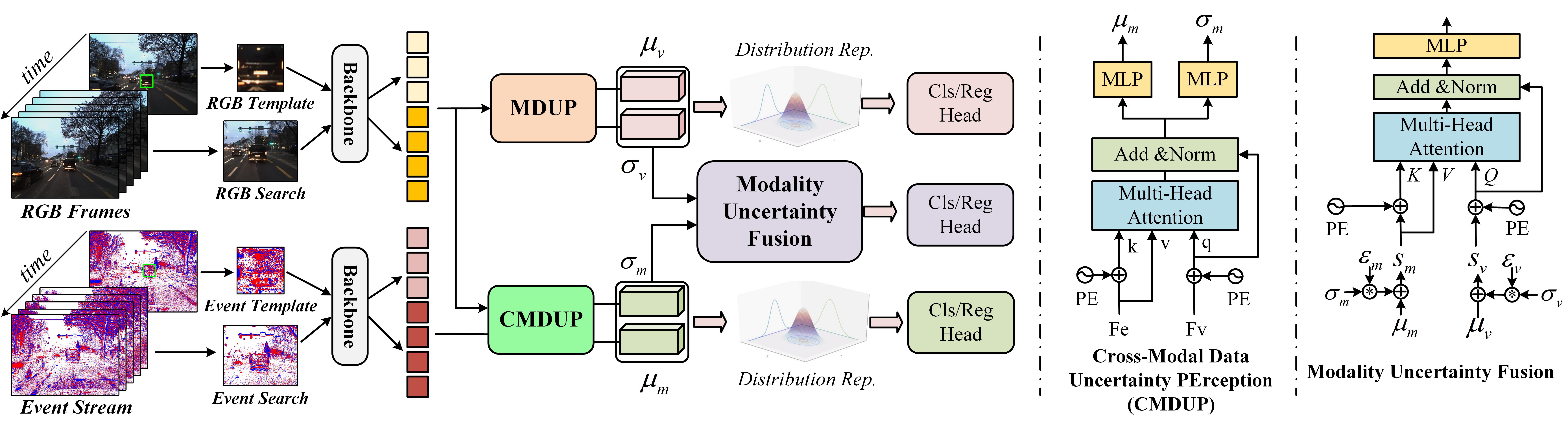}
\caption{An overview of our proposed framework for cross-resolution RGB-DVS based object tracking.} 
\label{framework}
\end{figure*}

\section{Methodology} \label{methodology}

\subsection{Overview} 
As shown in Fig.~\ref{framework}, given the RGB frames and event streams, following existing event-based trackers~\cite{wang2021viseventbenchmark, zhang2021MEEvent, chen2020end, chamorro2020high, chen2019asynchronous}, we first transform the continuous event streams into event images (a.k.a. surface) by stacking event points within a fixed time interval. Then, we resize the event images to make the resolution the same with RGB frames. 
Following OSTrack~\cite{ye2022joint}, we adopt the ViT~\cite{dosovitskiy2020image} network with token elimination as a unified backbone for the feature extraction. For both modalities, we extract the template and search region and feed them into the backbone for feature extraction. It is worth noting that our tracker predicts the probabilistic representation instead of regular point representation to better handle the relaxed registration. Then, we feed the template/search regions of RGB and event into the CMDUP (Cross-Modal Data Uncertainty Perception) module for uncertainty perception by predicting its distribution representation via mean and variation. Here, the mean represents the intrinsic feature of the fused modality and the variance denotes the uncertainty regarding the mean. By using the mean and variance to determine the Gaussian distribution, we are able to convert the point representation of the modality into a probabilistic representation, which enhances the generalization of the network. Due to significant differences between modalities, fused features are not always reliable. Therefore, we use the MDUP (Modal Data Uncertainty Perception) module to perform probability modeling of RGB branches as supplementary information for CMDUP. More importantly, we propose MUF (Modality Uncertainty Fusion) which adaptively fuses RGB-Event feature representation. Finally, we feed the enhanced features into a tracking head which contains both classification and regression branches for target object localization. In the following paragraphs, we will dive into the details of these modules.

\subsection{Input Encoding}  
Given the RGB frames $\mathcal{F}$ and event streams $\mathcal{E}$, we first stack the event streams into an image-like representation and resize its resolution to be the same as the RGB frames. Then, we extract the search and template regions of both modalities and divide them into image patches. Patch embedding layers are used to embed the input into token representations. Here, we denote the search and template tokens of RGB and Event data as $\mathcal{S}_v$, $\mathcal{T}_v$ and $\mathcal{S}_e$, $\mathcal{T}_e$. Then, we concatenate and feed RGB and Event tokens into the unified ViT~\cite{dosovitskiy2020image} backbone with token elimination proposed in OSTrack~\cite{ye2022joint}. More details about the backbone are referred to check their paper. 
\subsection{Cross-Modality Data Uncertainty Perception}
After we obtain the initial token representations from the backbone network (i.e., $\mathcal{S}'_v$, $\mathcal{T}'_v$ for RGB data, $\mathcal{S}'_e$, $\mathcal{T}'_e$ for event data), we will further process these features by considering the modality relations. Compared to the RGB frame, event data contains a large amount of noisy information and is also spatially sparse. Also, the RGB and Event data are not perfectly aligned which makes it a challenging task to fuse the dual modalities from the point of view of precise feature learning. Instead, we propose a cross-modal uncertainty estimation module to fuse the dual modalities which will be more robust for the unaligned RGB-Event tracking. Specifically, we feed the RGB tokens [$\mathcal{S}'_v$, $\mathcal{T}'_v$] into an MDUP (Modality Data Uncertainty PErception) module, and feed joint RGB-Event tokens [$\mathcal{S}'_v$, $\mathcal{T}'_v$, $\mathcal{S}'_e$, $\mathcal{T}'_e$] into the CMDUP (Cross-Modal Data Uncertainty PErception).

As shown in the right part of Fig.~\ref{framework}, CMDUP is a cross-attention style network that takes the RGB and Event tokens as the input, i.e., $F_v$ = [$\mathcal{S}'_v$, $\mathcal{T}'_v$], $F_e$ = [$\mathcal{S}'_e$, $\mathcal{T}'_e$]. The motivation for the selection of cross-attention is that it can effectively aggregate information from both modalities without relying on modality alignment. We project $F_v$ and $F_e$ into the query feature $\mathbf{Q}$, and key $\mathbf{K}$, value $\mathbf{V}$, respectively. In this procedure, the position encoding is also introduced and added to the token features. 
Mathematically, the computation of our CMDUP module can be written as: 
\begin{equation}
\begin{aligned}
&\operatorname{mAtt}(\mathbf{Q}, \mathbf{K}, \mathbf{V})=\left(\operatorname{Cat}\left(Head^1, \ldots, Head^N\right)\right) \mathbf{W}_o \\
&Head^j=\operatorname{Att}\left(\mathbf{Q} \mathbf{W}_1^j, \mathbf{K W}_2^j, \mathbf{V W}_3^j\right)\\
&\operatorname{Att}(\mathbf{Q}, \mathbf{K}, \mathbf{V})=\operatorname{Softmax}\left(\frac{\mathbf{Q} \mathbf{K}^{\mathrm{T}}}{\sqrt{c}}\right) \mathbf{V}
\end{aligned}
\end{equation}
where $\operatorname{Cat}$ denotes the concatenate operation, $\mathbf{W}_o \in \mathbb{R}^{C \times C}$, 
$\mathbf{W}_1^j \in \mathbb{R}^{C \times D}$, $\mathbf{W}_2^j \in \mathbb{R}^{C \times D}$, and $\mathbf{W}_3^j \in \mathbb{R}^{C \times D}$ are all learnable parameters. $D=C / N$, $N$ is the number of parallel attention heads. Then, two Multi-Layer Perceptron (MLP) are used to predict the mean $\mu$ and variance $\sigma$ of the fused features.

By obtaining the mean $\mu$ and variance $\sigma$ of the fused features, we can determine a Gaussian distribution, $\mathrm{p}\left(\mathbf{z}_{\mathrm{i}} \mid \mathbf{x}_{\mathrm{i}}\right)$. Specifically, we define the latent space representation $\mathbf{z}_{\mathrm{i}}$ of each sample $\mathbf{x}_{\mathrm{i}}$ as a Gaussian distribution, 
\begin{equation}
\mathrm{p}\left(\mathbf{z}_{\mathrm{i}} \mid \mathbf{x}_{\mathrm{i}}\right)=\mathcal{N}\left(\mathbf{z}_{\mathrm{i}} ; \mu_{\mathrm{i}} \sigma_{\mathrm{i}}^2 \mathbf{I}\right)
\end{equation}
where $\mathbf{I}$ represents the identity matrix which is a square matrix with diagonal elements equal to 1 and all other elements equal to 0. 
Note that, the representation of each feature is no longer a deterministic point embedding, but a random embedding sampled from the latent probability space, 
$\mathcal{N}\left(\mathbf{z}_{\mathrm{i}}; \mu_{\mathrm{i}} \sigma_{\mathrm{i}}^2 \mathbf{I}\right)$, 
to enhance the generalization ability of the network and better deal with noisy information. 
As the random sampling operation is not differentiable during model training, this will hinder the backward propagation of gradients. In this work, we employ the reparameterization technique~\cite{kingma2013auto} to enable the model to still take gradients as usual. Specifically, we first sample random noise from a normal distribution independent of the model parameters and then generate $\mathbf{s}_{\mathrm{i}}$ as an equivalent sampling representation: 
\begin{equation}
\mathbf{s}_{\mathrm{m}}=\mu_{\mathrm{m}}+\epsilon \sigma_{\mathrm{m}}, \epsilon \in \mathcal{N}(\mathbf{0}, \mathbf{I}). 
\end{equation}

For the RGB modality, we adopt a similar procedure by proposing MDUP to enhance the RGB feature learning, and generate equivalent sample $\mathbf{S}_{\mathrm{v}}$. The difference with the CMDUP module is that the input of this module is RGB tokens only. Then, we take the sampled embeddings and feed them into the tracking head for target object localization. 

\subsection{Modality Uncertainty Fusion} 
To achieve more robust tracking results, in this work, we propose a modality uncertainty fusion module to fuse the RGB and Event representations effectively. As shown in the right part of Fig.~\ref{framework}, given the two equivalent samples $\mathbf{S}_{\mathrm{m}}$ and  $\mathbf{S}_{\mathrm{v}}$ from RGB and Event branch, the $\mathbf{S}_{\mathrm{v}}$ and $\mathbf{S}_{\mathrm{m}}$ are used as the query $Q$ and the key $K$, value $V$, respectively. A cross-attention block which contains multi-head attention layers is used to fuse these inputs and an MLP layer is adopted to get the final features.

\subsection{Loss Function} 
In the training phase, all embedding $\mu_{\mathrm{i}}$ are disrupted by $\sigma_{\mathrm{i}}$. This encourages the model to predict small $\sigma$ for all samples to suppress uncertain components in $s_{\mathrm{i}}$, ensuring convergence of the network. In this case, the random representation can be rewritten as $\mathbf{s}_{\mathrm{i}}=\mu_{\mathrm{i}}+\mathbf{c}$, which effectively degenerates into the original deterministic representation. Inspired by variational information bottleneck, we introduce a regularization term in the optimization process which can explicitly constrain the distribution $\mathcal{N}\left(\mu_{\mathrm{i}}, \sigma_{\mathrm{i}}\right)$ to be close to the normal distribution $\mathcal{N}(\mathbf{0}, \mathbf{I})$ by measuring the Kullback-Leibler divergence (KLD) between the two distributions. The KLD can be formulated as follows:
\begin{equation}
\begin{aligned}
\mathcal{L}^{kl} & =K L\left[N\left(\mathbf{z}_i \mid \boldsymbol{\mu}_i, \boldsymbol{\sigma}_i^2\right)|| N(\epsilon \mid \mathbf{0}, \mathbf{I})\right] \\
& =-\frac{1}{2}\left(1+\log \boldsymbol{\sigma}^2-\boldsymbol{\mu}^2-\boldsymbol{\sigma}^2\right)
\end{aligned}
\end{equation}
Here, we model the data uncertainty for both the RGB branch and the cross-modal branch separately and denote the regularization term for the loss of each branch as $\mathcal{L}_{v}^{kl}$ and $\mathcal{L}_{cm}^{kl}$, respectively.

Following OSTrack~\cite{ye2022joint}, we employ the weighted focal loss~\cite{law2018cornernet} for classification, the $\ell_1$ loss and the generalized IoU loss~\cite{rezatofighi2019generalized} for bounding box regression. The loss functions used in the cross-modal branch can be represented as: 
\begin{equation}
\mathcal{L}_{\text {cm}}=\mathcal{L}_{\text {cls}}+\lambda_{\text {iou }} \mathcal{L}_{\text {iou }}+\lambda_{\mathcal{L}_1} \mathcal{L}_1
\end{equation}
where $\lambda_{\text {iou }}$ and $\lambda_{\mathcal{L}_1}$ are weight factors and are set to 2 and 5, respectively. Similarly, the classification and regression losses for the final fusion branch and RGB branch can be represented as $\mathcal{L}_{\text {f}}$ and $\mathcal{L}_{\text {v}}$, respectively. 
Therefore, the overall loss functions can be written as: 
\begin{equation}
\mathcal{L}_{total}= \mathcal{L}_{\text {f}}+\mathcal{L}_{\text {cm}}+\mathcal{L}_{\text {v}}+\alpha(\mathcal{L}_{v}^{kl}+\mathcal{L}_{cm}^{kl})
\end{equation}
where $\alpha$ is set to 0.001.

\section{Experiment} \label{experiment}

\subsection{Dataset and Evaluation Metric} 
In this paper, we conduct extensive experiments on three RGB-DVS tracking datasets, including \textbf{VisEvent}~\cite{wang2021viseventbenchmark}, \textbf{COESOT}~\cite{tang2022COESOT}, and our newly proposed \textbf{CRSOT}. We train our tracker on the training subset and evaluate the results on the corresponding testing subset of these datasets. For the evaluation, we adopt the popular One-Pass Evaluation (OPE) by following OTB benchmark~\cite{wu2015OTB} and report the results of \textbf{Precision Rate (PR)}, \textbf{Success Rate (SR)}, and \textbf{Normalized Precision Rate (NPR)}. 

\subsection{Implementation Details} 
In the training phase, we set the learning rate of the backbone to 0.000005 and set the learning rate of other parameters to 0.00005. The weight decay is 0.0001 and a decay factor of 0.2 is employed after 50 epochs. We adopt the AdamW~\cite{ilya2019decoupled} to optimize our network. To ensure fairness, we strictly follow the settings of other algorithms during training. We train our tracker on the training subset of CRSOT, COESOT, and VisEvent for 20, 10, and 30 epochs, respectively.

\begin{table}
\centering
\small 
\caption{\textbf{PR, NPR, and SR scores (\%) of our tracker on CRSOT dataset against other trackers.} 
The best results are highlighted in $\color{red} \textbf{red}$ color. 
* indicates that the tracker is re-trained using the CRSOT training dataset.}
\begin{tabular}{c|l|ccc}
\hline
\textbf{Input} &\textbf{Methods}  &  \multicolumn{3}{c}{\textbf{CRSOT}}  \\
& & PR$\uparrow$ & NPR$\uparrow$   & SR$\uparrow$ \\
\hline
&ATOM~\cite{danelljan2019atom} &62.9 & 64.0 & 50.5 \\	
&DiMP50~\cite{bhat2019dimp} & 62.8 & 64.3 & 52.1 \\	
&PrDiMP50~\cite{prdimp} &61.2 & 63.0 & 51.6 \\
&Super\_DiMP & 63.7 & 65.5 & 53.7 \\
\textbf{RGB} &Keep\_Track~\cite{keeptrack} & 64.4 & 66.2 & 54.0 \\
&TransT~\cite{transt}  & 65.5 & 65.9 & 54.0 \\
&Trdimp~\cite{trdimp}  & 65.3 & 66.3 & 54.4 \\
&ToMP50~\cite{tomp} & 63.7 & 63.9 & 52.9 \\
&ToMP50*~\cite{tomp} & 69.6 & 71.1 & 59.0 \\
\hline
&DiMP50~\cite{bhat2019dimp}& 52.3 & 54.7 & 43.3 \\
&DiMP50*~\cite{bhat2019dimp} & 65.8 & 67.7 & 54.8 \\
&ToMP50~\cite{tomp} & 53.2 & 53.8 & 44.5 \\
&ToMP50*~\cite{tomp} & 63.9 & 66.6 & 54.6 \\
&Keep\_Track~\cite{keeptrack} & 53.5 & 55.5 & 43.8 \\
&MixFormer*~\cite{mixformer} & 63.6 & 64.5 &53.3 \\
&SeqTrack~\cite{chen2023seqtrack} & 58.5 & 59.5 &48.3 \\
\textbf{RGB-DVS} &GRM~\cite{gao2023generalized} & 59.7 & 61.1 &50.5 \\
&GRM*~\cite{gao2023generalized} & 49.8 & 51.0 &42.3 \\
&ROMTrack~\cite{cai2023robust} & 60.8 & 62.4 &51.2 \\
&ARTrack~\cite{wei2023autoregressive} & 61.6 & 63.1 &52.5 \\
&ARTrack*~\cite{wei2023autoregressive} & 68.1 & 69.3 &56.8 \\
&OSTrack*~\cite{ye2022joint} & 66.1 & 67.5 &55.5 \\
&ViPT*~\cite{zhu2023visual} & 64.9 & 66.0 &54.6 \\
&Ours & \color{red}\textbf{74.2} & \color{red}\textbf{74.4} & \color{red}\textbf{61.8} \\
\hline
\end{tabular}
\label{CRSOT_Result}
\end{table}

\subsection{Comparison on Public Benchmarks}  
In this section, we report and compare our tracking results on three RGB-Event based tracking benchmark datasets, including CRSOT, VisEvent, and COESOT. 
For the CRSOT dataset, as shown in Table~\ref{CRSOT_Result}, our baseline OSTrack~\cite{ye2022joint} achieves $66.1/67.5/55.5$ on the PR/NPR/SR metric, respectively. When introducing the uncertainty-aware feature learning module, our results are $74.2/74.4/61.8$ on these metrics, which fully validated the effectiveness of our proposed modules for not strictly aligned RGB-DVS tracking. When compared with other SOTA trackers, like DiMP50~\cite{bhat2019dimp}, MixFormer~\cite{mixformer}, SeqTrack~\cite{chen2023seqtrack}, our tracking results are also better than theirs which are new state-of-the-art on the CRSOT benchmark dataset.

\begin{figure}
\center
\includegraphics[width=\columnwidth]{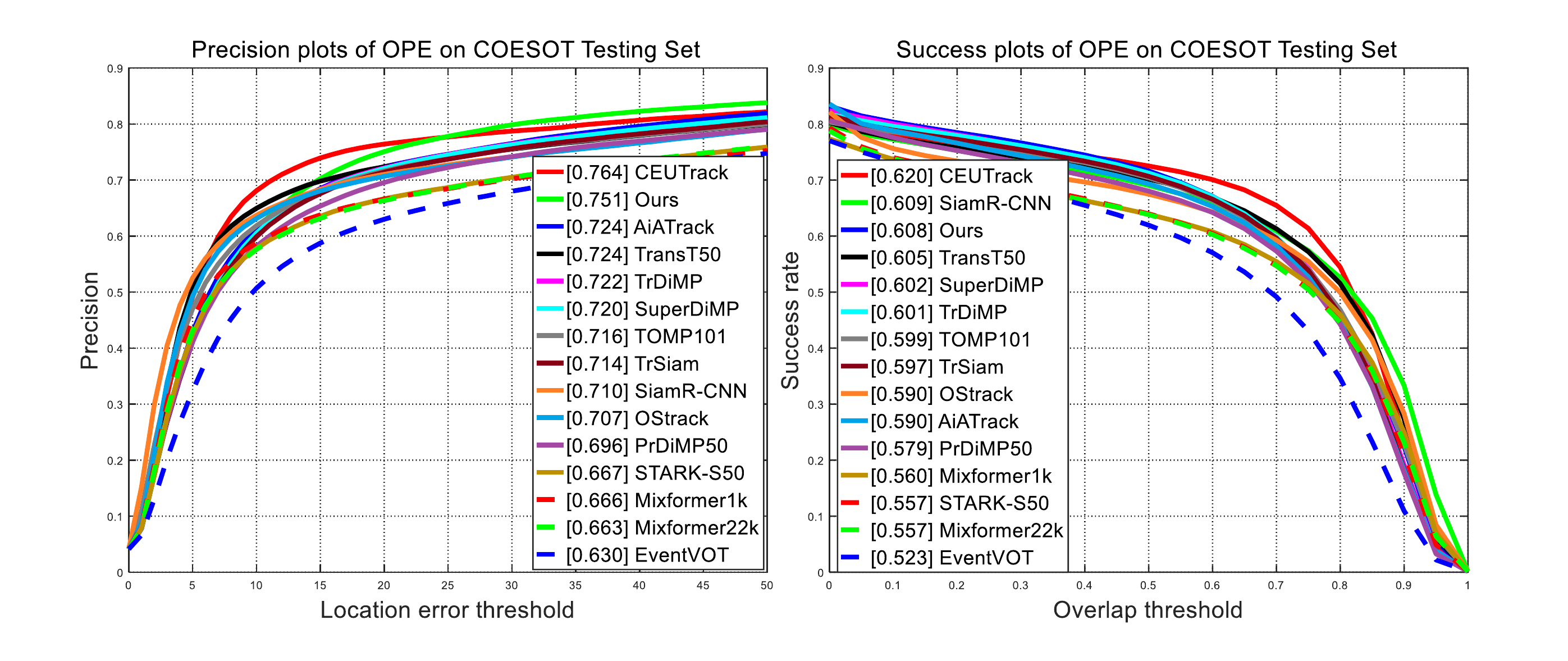}
\caption{Tracking results of our tracker and other state-of-the-art trackers on COESOT testing set.} 
\label{fig:COESOT} 
\end{figure}

For the COESOT dataset, as shown in Fig.~\ref{fig:COESOT}, we can find that our tracker achieves second and third place on the PR and SR metrics ($0.751/0.608$), respectively. On the VisEvent dataset, as illustrated in Table~\ref{viseventresults}, we obtain 52.5/74.1 on the SR/PR metric which is also better than most of the compared strong trackers. Therefore, we can draw the conclusion that our tracker achieves state-of-the-art performance on existing and newly proposed frame-event tracking datasets.

\begin{table*}
\centering
\caption{Experimental results on VisEvent testing set.} 
\label{viseventresults}
\resizebox{\textwidth}{!}{
\begin{tabular}{c|lllllllllll}
\toprule
Tracker & \multicolumn{1}{c}{ATOM(EF)~\cite{danelljan2019atom}}& \multicolumn{1}{c}{DiMP50(EF)~\cite{dimp}}  & \multicolumn{1}{c}{ProTrack~\cite{yang2022prompting}}& \multicolumn{1}{c}{PrDiMP50(EF)~\cite{prdimp}} &
\multicolumn{1}{c}{OSTrack~\cite{ye2022joint}} &
\multicolumn{1}{c}{STARKS50~\cite{stark}}
& \multicolumn{1}{c}{SiamBAN(EF)~\cite{siamban}} 
& \multicolumn{1}{c}{MDNet(MF)~\cite{nam2016mdnet}}  
& \multicolumn{1}{c}{ SiamRCNN(EF)~\cite{voigtlaender2020siam}}    
   & \multicolumn{1}{c}{ViPT~\cite{zhu2023visual}} & \multicolumn{1}{c}{\textbf{Ours}}  \\
\midrule
SR  & \multicolumn{1}{c}{41.2} & \multicolumn{1}{c}{45.1}   & \multicolumn{1}{c}{47.1}   & \multicolumn{1}{c}{45.3}  & \multicolumn{1}{c}{53.4} &\multicolumn{1}{c}{44.6}  & \multicolumn{1}{c}{40.5} &\multicolumn{1}{c}{42.6} &\multicolumn{1}{c}{49.9} &\multicolumn{1}{c}{59.2}  & \multicolumn{1}{c}{\textbf{52.5}}\\
PR  & \multicolumn{1}{c}{60.8} & \multicolumn{1}{c}{66.1}   & \multicolumn{1}{c}{63.2}   & \multicolumn{1}{c}{64.4}  & \multicolumn{1}{c}{69.5}  & \multicolumn{1}{c}{61.2}  &\multicolumn{1}{c}{59.1}  &\multicolumn{1}{c}{66.1}  &\multicolumn{1}{c}{65.9} 
 &\multicolumn{1}{c}{75.8} & \multicolumn{1}{c}{\textbf{74.1}}\\
\bottomrule
\end{tabular}
}
\end{table*}


\begin{table}
\setlength{\tabcolsep}{0.17cm}
\centering
\small 
\caption{Component analysis of our proposed framework on the CRSOT dataset.}  
\label{CompAnalysis} 
\begin{tabular}{cccc|ccc}
\hline 
\textbf{Baseline}  &\textbf{ MDUP }  & \textbf{CMDUP} & \textbf{MUF}& \textbf{PR} & \textbf{NPR} & \textbf{SR} \\
\hline 
\cmark&\xmark &\xmark &\xmark & 71.9 & 72.5 & 60.3 \\
\rowcolor{mygray}
\cmark&\cmark &\xmark &\xmark & 72.8 & 73.3 & 61.3 \\
\cmark&\xmark &\cmark &\xmark & 73.1 & 73.2 & 61.2 \\
\cmark&\cmark &\cmark &\xmark & 73.5 & 73.6 & 61.6 \\
\rowcolor{mygray}
\cmark&\cmark &\cmark &\cmark & \bf 74.2 & \bf 74.4 & \bf 61.8 \\     
\hline
\end{tabular}
\end{table}

\subsection{Ablation Study}   

\noindent \textbf{Component Analysis.~} 
There are three key components in our proposed frame-event tracking framework, including MDUP, CMDUP, and MUF. As illustrated in Table~\ref{CompAnalysis}, our baseline achieves $71.9/72.5/60.3$ on the PR/NPR/SR metrics on the CRSOT dataset, respectively. Note that compared to directly adding two modalities as input, we use a $1\times1$ convolution to concatenate the information of the two modalities as the input to the baseline, which has stronger performance. When introducing the MDUP, the results can be improved to $72.8/73.3/61.3$. If we utilize the CMDUP based on the baseline tracker, the results can also be boosted to $73.1/73.2/61.2$. When both MDUP and CMDUP are used, the results are $73.5/73.6/61.6$. When all three modules are used, the best tracking results can be obtained, i.e., $74.2/74.4/61.8$. From these experimental results and analysis, we can find that all our proposed modules contribute to the final tracking results.

\begin{figure}
\center
\includegraphics[width=0.8\columnwidth]{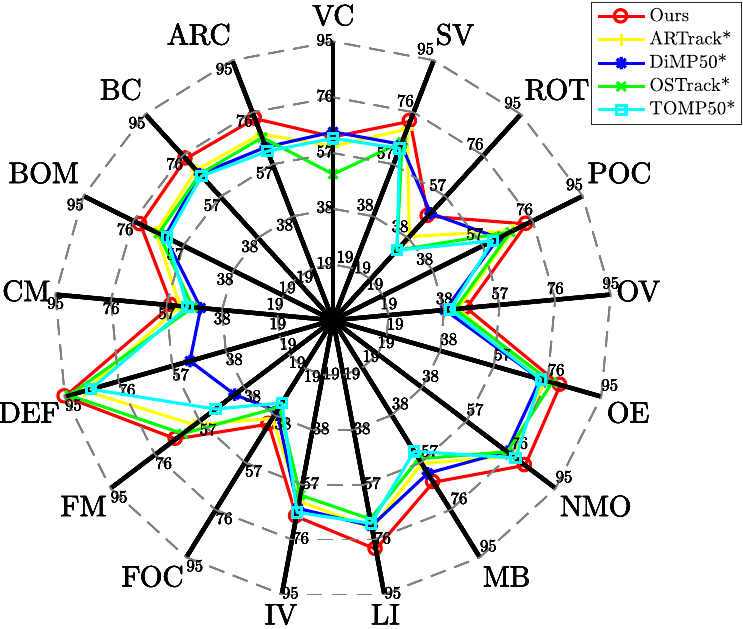}
\caption{NPR of different attributes on the CRSOT dataset.} 
\label{attributeResults} 
\end{figure}

\noindent \textbf{Attribute Analysis.~}  
In our proposed CRSOT dataset, 17 attributes are defined based on features of unimodal and bimodal data. 
As shown in Fig.~\ref{attributeResults}, our tracking results are significantly better than the compared trackers, including ARTrack, DiMP50, OSTrack, and TOMP50. It is also easy to find that current trackers perform well on DEF, however, these trackers perform poorly on CM, OV, and FOC. These experiments demonstrate that the RGB-Event-based tracking is far from addressed well.

\noindent \textbf{Efficiency Analysis and Model Parameters.~} 
Our proposed tracker achieves 32 FPS on the CRSOT dataset. The scale of our tracking model is 470.2 MB, and it contains 117.5 MB parameters.

\subsection{Visualization} 

\begin{figure}
\center
\includegraphics[width=\columnwidth]{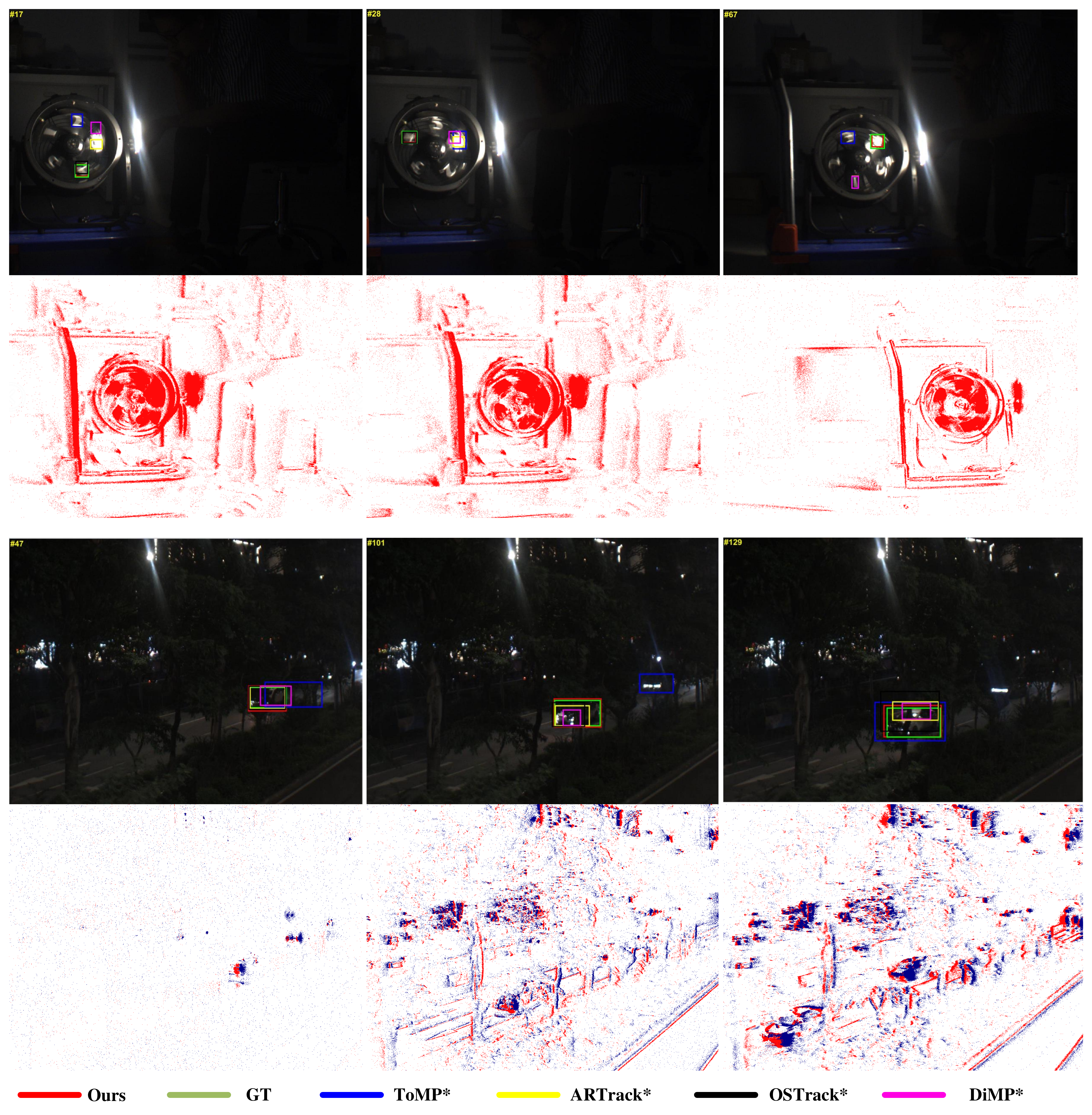}
\caption{Tracking results of ours and other SOTA trackers.}  
\label{TrackresultsVIS} 
\end{figure}

In this section, we provide some visualizations of our tracking results to further help the readers understand our proposed tracker. As shown in Fig.~\ref{TrackresultsVIS}, it's hard for visual trackers to track in low illumination scenarios, meanwhile, the event streams provide good supplementary information which makes trackers achieve a higher tracking performance. Our tracking results are more robust than the compared trackers as illustrated in this visualization.

\subsection{Limitation Analysis} 
Although our proposed tracking algorithm achieves a higher tracking performance on multiple benchmark datasets, however, our tracker still can be further enhanced from the following aspects: 
1). The encoding of event streams can be replaced using spiking neural networks to achieve energy-efficient feature learning; 
2). As the RGB-Event video pairs are not perfectly aligned, how to learn features from such roughly aligned videos is worth designing new alignment modules to try to solve. 
We will these as our future works.

\section{Conclusion} \label{conclusion}
Tracking using RGB and event cameras has drawn more and more attention in recent years, however, existing RGB-Event tracking datasets are collected using DVS346 with limited resolutions. In this work, we formally propose a new task single object tracking which fuses the unaligned neuromorphic and visible cameras, and propose a new dataset which is collected using high-resolution RGB and Event cameras. We build the first unaligned frame-event dataset CRSOT collected with a specially built data acquisition system, which contains 1,030 high-definition RGB-Event video pairs, 304,974 video frames. In addition, we also propose a new baseline approach that models the RGB-Event feature fusion using uncertain-aware learning. Extensive experiments demonstrate that our tracker can collaborate the dual modalities for high-performance tracking without strictly temporal and spatial alignment. In our future works, we will consider designing low-latency and energy-efficient backbones for the unaligned frame-event single object tracking.

{
    \small
    \bibliographystyle{ieeenat_fullname}
    \bibliography{reference}
}

\end{document}